\documentclass[sigconf]{acmart}

\usepackage{graphicx}
\usepackage{amsmath}
\usepackage{epstopdf}
\usepackage{array}
\usepackage{multirow}
\usepackage{amsfonts}
\usepackage{booktabs}
\usepackage{subfigure}
\usepackage{bm}
\usepackage{algorithm}
\usepackage{algpseudocode}
\usepackage{color}
\usepackage{multirow}

\AtBeginDocument{%
  \providecommand\BibTeX{{%
	\normalfont B\kern-0.5em{\scshape i\kern-0.25em b}\kern-0.8em\TeX}}}

\copyrightyear{2020}
\acmYear{2020}
\setcopyright{acmcopyright}\acmConference[MM '20]{Proceedings of the 28th ACM International Conference on Multimedia}{October 12--16, 2020}{Seattle, WA, USA}
\acmBooktitle{Proceedings of the 28th ACM International Conference on Multimedia (MM '20), October 12--16, 2020, Seattle, WA, USA}
\acmPrice{15.00}
\acmDOI{10.1145/3394171.3413822}
\acmISBN{978-1-4503-7988-5/20/10}

\begin{document}

\title{Adversarial Graph Representation Adaptation for Cross-Domain Facial Expression Recognition}

\author{Yuan Xie}
\email{phoenixsysu@gmail.com}
\affiliation{%
  \institution{DarkMatter Research}
  \city{Guangzhou}
  \country{China}
}

\author{Tianshui Chen}
\authornote{Tianshui Chen is the corresponding author.}
\email{tianshuichen@gmail.com}
\affiliation{%
  \institution{DarkMatter Research}
  \institution{Sun Yat-Sen University}
  \city{Guangzhou}
  \country{China}
}

\author{Tao Pu}
\email{putao537@gmail.com}
\affiliation{%
  \institution{Sun Yat-Sen University}
  \city{Guangzhou}
  \country{China}
}

\author{Hefeng Wu}
\email{wuhefeng@gmail.com}
\affiliation{%
 \institution{Sun Yat-Sen University}
 \city{Guangzhou}
  \country{China}
 }

\author{Liang Lin}
\affiliation{%
  \institution{Sun Yat-Sen University}
  \institution{DarkMatter Research}
  \city{Guangzhou}
  \country{China}
}

\renewcommand{\shortauthors}{Yuan Xie, et al.}

\begin{abstract}
 Data inconsistency and bias are inevitable among different facial expression recognition (FER) datasets due to subjective annotating process and different collecting conditions. Recent works resort to adversarial mechanisms that learn domain-invariant features to mitigate domain shift. However, most of these works focus on holistic feature adaptation, and they ignore local features that are more transferable across different datasets. Moreover, local features carry more detailed and discriminative content for expression recognition, and thus integrating local features may enable fine-grained adaptation. In this work, we propose a novel Adversarial Graph Representation Adaptation (AGRA) framework that unifies graph representation propagation with adversarial learning for cross-domain holistic-local feature co-adaptation. To achieve this, we first build a graph to correlate holistic and local regions within each domain and another graph to correlate these regions across different domains. Then, we learn the per-class statistical distribution of each domain and extract holistic-local features from the input image to initialize the corresponding graph nodes. Finally, we introduce two stacked graph convolution networks to propagate holistic-local feature within each domain to explore their interaction and across different domains for holistic-local feature co-adaptation. In this way, the AGRA framework can adaptively learn fine-grained domain-invariant features and thus facilitate cross-domain expression recognition. We conduct extensive and fair experiments on several popular benchmarks and show that the proposed AGRA framework achieves superior performance over previous state-of-the-art methods.
\end{abstract}

\ccsdesc[500]{Computing methodologies~Computer vision problems}
\ccsdesc[500]{Computing methodologies~Image representations}
\ccsdesc[500]{Computing methodologies~Cross-validation}

\keywords{Facial expression recognition; Domain adaptation; Graph neural network}


\maketitle

\section{Introduction}
Facial expressions carry human emotional states and intentions. Thus, automatically recognizing facial expressions helps understand human behaviors, which benefits a wide range of applications ranging from human-computer interaction \cite{fragopanagos2005emotion} to medicine \cite{edwards2002emotion} and security monitoring \cite{clavel2008fear,saste2017emotion}. During the last decade, lots of efforts are dedicated to collect variant FER datasets, including lab-controlled (e.g., CK+ \cite{lucey2010extended}, JAFFE \cite{lyons1998coding}, MMI \cite{valstar2010induced}, Oulu-CASIA \cite{zhao2011facial}) and in-the-wild (e.g., RAF \cite{li2017reliable,li2018reliable}, SFEW2.0 \cite{dhall2011static}, ExpW \cite{zhang2015learning}, FER2013 \cite{goodfellow2015challenges}) environments. With these datasets, variant deep learning models are intensively proposed, progressively facilitating the FER performance.

However, because human's understanding of facial expressions varies with their experiences and living cultures, their annotations are inevitably subjective, leading to obvious domain shifts across different datasets \cite{Zeng_2018_ECCV,li2020deeper}. In addition, facial images of different datasets are usually collected in different environments (e.g., lab-controlled or in-the-wild) and from humans of different races, which further enlarge this domain shift \cite{li2018deep}. Consequently, current best-performing methods may achieve satisfactory performance in within-dataset protocols, but they suffer from dramatic performance deterioration in cross-dataset settings \cite{Zeng_2018_ECCV}. To mitigate the domain shift, recent works \cite{tzeng2017adversarial,long2018conditional,wang2018deep,goodfellow2014generative} introduce adversarial learning mechanisms that aim to learn domain-invariant features, and some researchers also adapt these mechanisms to cross-domain FER \cite{wei2018unsupervised,wang2018unsupervised}. Despite achieving acknowledged progress, these works focus on extracting holistic features for domain adaptation, and they ignore local features that benefit cross-domain FER from the two aspects: i) Local regions carry discriminative features that are more transferable across different datasets. For example, the lip-corner-puller action is discriminative to distinguish happy expression and it is similar for samples of different datasets. ii) Local regions encode more detailed and complementary features to holistic features. Modeling the correlation of holistic-local features within each domain and across different domains may enable finer-grained adaptation and thus facilitate cross-domain FER.

\begin{figure}[!t]
\centering
{
\label{fig:subfig1} 
\includegraphics[width=0.48\linewidth]{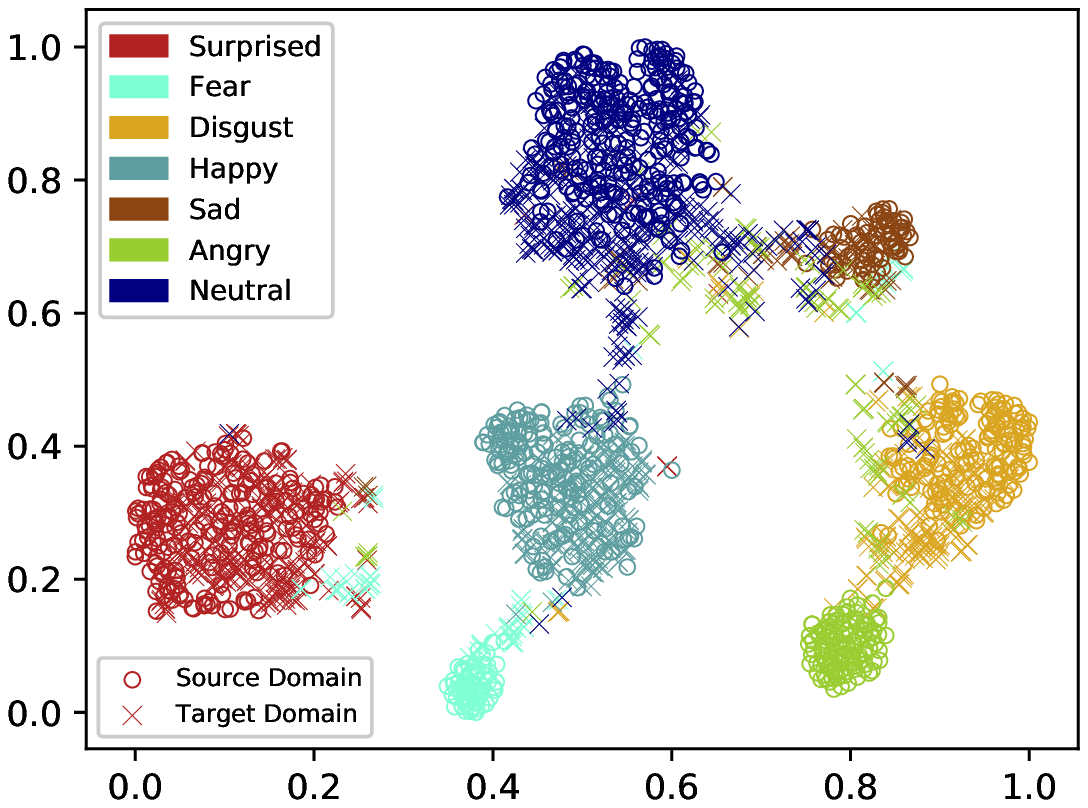}}{
\label{fig:subfig2} 
\includegraphics[width=0.48\linewidth]{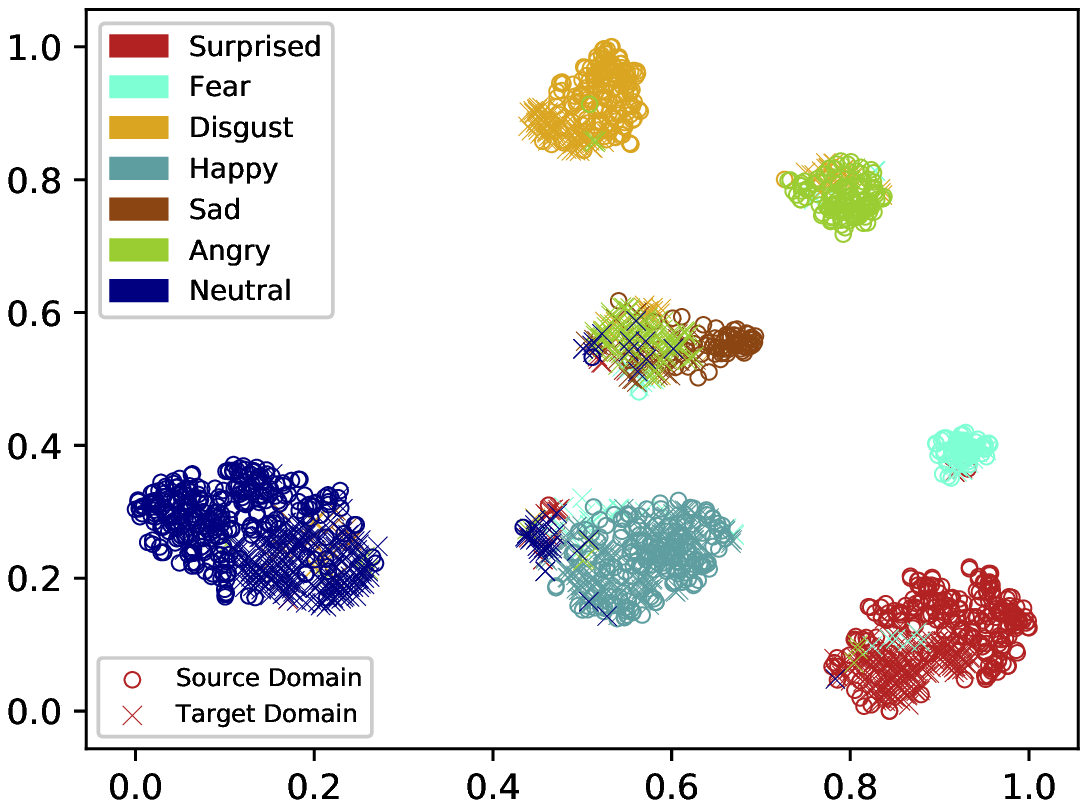}}
\vspace{-10pt}
\caption{Illustration of feature distribution learned by the baseline adversarial learning \cite{tzeng2017adversarial} method that merely uses holistic features (left) and our proposed AGRA framework. It is obvious that the AGRA framework can better gather the samples of the same category and from different domains together, suggesting it can learn more discriminative domain-invariant features for cross-domain FER.}
\vspace{-20pt}
\label{fig:motivation}
\end{figure}

 In this work, we show that the correlation of holistic-local features within each domain and across the source and target domains can be explicitly represented by structured graphs, and their interplay and adaptation can be captured by adaptive message propagation through the graphs. To achieve this, we develop a novel Adversarial Graph Representation Adaptation (AGRA) framework, which integrates graph representation propagation with adversarial learning mechanism for cross-domain holistic-local feature interplay and co-adaptation. Specifically, we first extract several discriminative local regions based on facial landmarks (e.g., eyes, nose, mouth corner, etc.) \cite{zhang2016joint,liu2019facial} and build two graphs to correlate holistic image and local regions within each domain and across different domains, respectively. Given an input image from one domain, we extract features of the holistic image and the local regions to initialize the corresponding nodes of this domain. The nodes of the other domain are initialized by the corresponding per-class learnable statistical feature distributions. Then, we introduce two stacked graph convolution networks to propagate node messages within each domain to explore holistic-local features interaction and across the two different domains to enable holistic-local feature co-adaptation. In this way, it can progressively mitigate the shift of the holistic-local features between the source and target domains, enabling learning discriminative and domain-invariant features to facilitate cross-domain FER. Figure \ref{fig:motivation} shows the feature distributions learned by the baseline adversarial learning \cite{tzeng2017adversarial} method that merely uses holistic features for domain adaption and our proposed AGRA framework. It shows that our framework can better gather the features of samples that belong to the same category and are taken from different domains together, suggesting that it can better learn domain-invariant features while improving their discriminative ability.

In summary, the contributions of this work are three-fold. First, we propose to integrate graph representation propagation with the adversarial learning mechanism for holistic-local feature co-adaptation across different domains. It enables learning more discriminative and domain-invariant features for cross-domain FER. Second, we develop a class-aware two-stage updating mechanism to iteratively learn the statistical feature distribution of each domain for graph node initialization. It plays a key role to mitigate domain shift that facilitates learning domain-invariant features. Finally, we conduct extensive and fair experiments on public benchmarks to demonstrate the effectiveness of the proposed framework. When using the same ResNet-50 as the backbone and the same RAF-DB \cite{li2018reliable} as the source dataset, the proposed framework improves the accuracy averaging over the CK+ \cite{lucey2010extended}, JAFFE \cite{lyons1998coding}, FER2013 \cite{goodfellow2015challenges}, SFEW2.0 \cite{dhall2011static}, and ExpW \cite{zhang2018facial} datasets by 4.02\% compared with the previous best-performing works. The codes and trained models are available at \url{http://github.com}.

\begin{figure*}[!t]
   \centering
   \includegraphics[width=0.8\linewidth]{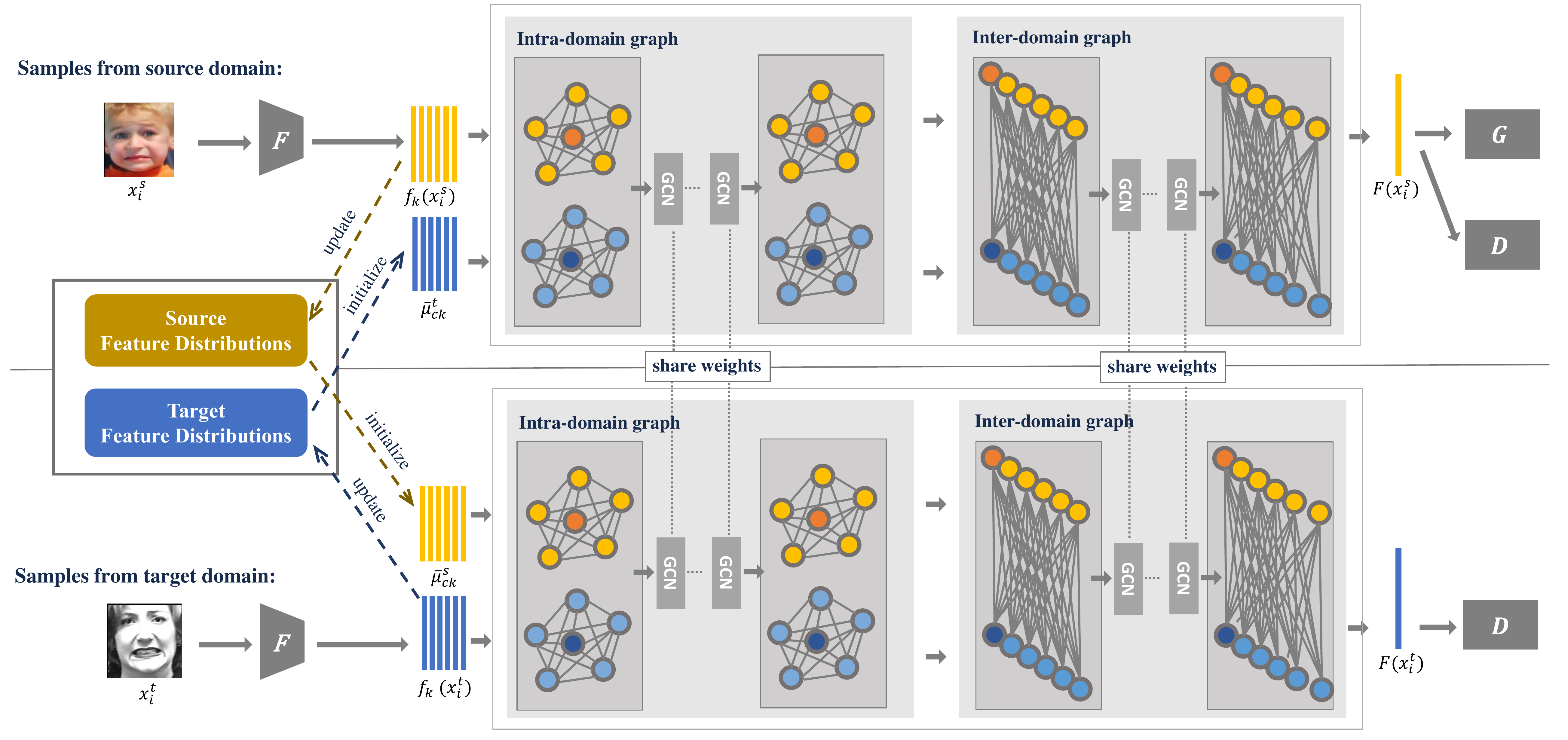}
   \vspace{-10pt}
   \caption{An illustration of the proposed Adversarial Graph Representation Adaptation framework. The framework builds two graphs to correlate holistic-local features within each domain and cross different domains respectively, initializes the graph nodes with input image features of a certain domain and the learnable statistical distribution of the other domain, and introduces two stacked GCNs to propagate node information within each domain and transfer node messages across different domains for holistic-local feature co-adaptation. Note that the nodes in the intra-domain and inter-domain graphs are the same, and we arrange them in different layouts for more clear connection illustration.}
   \Description[<short description>]{<long description>}
   \vspace{0pt}
   \label{fig:framework}
\end{figure*}

\section{Related Works}
In this section, we mainly review two stream of most-related works, i.e., cross-domain FER and adversarial domain adaptation.

\subsection{Cross-Domain FER}
Due to the subjective annotation and collection inconsistency, the distribution divergences inevitably exist among different cross-domain FER datasets. To deal with this issue, a series of works \cite{miao2012cross,sangineto2014we,yan2016transfer,zhu2016discriminative,yan2016cross,zheng2016cross,chu2016selective,li2020deeper,zong2018domain,yan2019cross} are dedicated to cross-domain FER. For example, Yan et al. \cite{yan2016transfer} propose a subspace learning method that transfer the knowledge obtained from the source dataset to the target dataset. However, this method still requires some annotated samples from the target dataset, which is unexpected in unsupervised scenarios. To facilitate cross-domain FER in an unsupervised manner, Zheng et al. \cite{zheng2016cross} further design a transductive transfer subspace learning method that combines the labeled source data and unlabeled auxiliary target data to jointly learn a discriminative subspace for cross-pose and cross-database FER. Different from these two works, Zong et al. \cite{zong2018domain} address this task by re-generating source and target samples that share the same or similar feature distribution. Wang et al. \cite{wang2018unsupervised} further introduce generative adversarial networks \cite{goodfellow2014generative} to generate samples of the target domain to fine-tune the model. More recently, Li et al. \cite{li2020deeper} observes that conditional probability distributions between source and target datasets are different. And they develop a deep emotion-conditional adaption network that simultaneously considers conditional distribution bias and expression class imbalance problem in cross-domain FER. In this work, we devise a new framework that integrates graph propagate networks with adversarial learning mechanisms for adaptive holistic-local feature co-adaptation, which learns fine-grained and domain-invariant features to facilitate cross-domain FER.

\subsection{Adversarial Domain Adaptation}
Domain discrepancies commonly exist across different datasets, and variant domain adaptation methods \cite{tzeng2017adversarial,long2018conditional,xu2019larger} are intensively proposed to learn domain-invariant feature thus that classier/predictor learned using the source datasets can be generalized to the target test datasets. Motivated by the generative adversarial networks \cite{goodfellow2014generative} that aims to generate samples that are indistinguishable from the real samples, recent domain adaptation works \cite{tzeng2017adversarial,long2018conditional} also resort to adversarial learning to mitigate domain shift. Specifically, it involves a two-player game, in which a feature extractor aims to learn transferable domain-invariant features while a domain discriminator struggles to distinguish samples from the source domain from those from the target domain. As a pioneer work, Tzeng et al. \cite{tzeng2017adversarial} propose a generalized adversarial adaptation framework by combining discriminative modeling, untied weight sharing, and an adversarial loss. Long et al. further design a conditional adversarial domain adaptation method that further introduces two strategies of multi-linear conditioning and entropy conditioning to improve the discriminability and controls the uncertainty of classifier. Despite achieving impressive progress for cross-domain general image classification, these methods \cite{tzeng2017adversarial,long2018conditional} mainly focus on holistic features for adaptation and ignore local content that carries more transferable and discriminative features. Different from these works, we introduce graph propagation networks to capture the interplay of holistic-local features with each domain and across different domains, and integrate it with adversarial learning mechanisms to enable holistic-local feature co-adaptation. 

The proposed framework is also related to some works \cite{chen2019knowledge,wang2018deep1,chen2020knowledge,chen2019learning,chen2018knowledge,DBLP:conf/cvpr/ChenWWG19,wang2018zero} that adapts graph neural networks \cite{kipf2016semi,li2015gated} for visual interaction learning and reasoning. These works propose to explicitly model label dependencies in the form of graphs and adopt the graph to guide feature interaction learning. Inspired by these works, we further extend the graph to model within-domain and cross-domain holistic-local feature interactions that enables fine-grained feature adaption. 

\section{AGRA Framework}

\subsection{Overview}
We first introduce the cross-domain FER task, in which a source domain dataset $\mathcal{D}_s=\{(x^s_i, y^s_i)\}_{i=1}^{n_s}$ and a target domain dataset $\mathcal{D}_t=\{(x^t_i)\}_{j=1}^{n_t}$ are provided. The two datasets are sampled from two different distributions $p_s(X, Y)$ and $p_t(X, Y)$, respectively. Each sample from the source data $x^s_i$ has a label $y^s_i$ while the samples from the target dataset have no label. To address this task, the proposed AGRA framework builds on the adversarial cross-domain mechanism that learns domain-invariant features via a two-player games
\begin{eqnarray}
&&\mathop{\min}\limits_{D}  \mathcal{L}(F, G, D) \label{eqn:al1}\\
&&\mathop{\min}\limits_{G, F} \mathcal{L}(F, G) - \mathcal{L}(F, G, D) \label{eqn:al2}
\end{eqnarray}
where
\begin{equation}
   \begin{split}
    \mathcal{L}(F, G)=&-\mathbb{E}_{(x^s, y^s)\sim \mathcal{D}_s}\ell(G(F(x^s)), y^s)\\
    \mathcal{L}(F, G, D)=&-\mathbb{E}_{(x^s, y^s)\sim \mathcal{D}_s}\log\left[D(F(x^s))\right]\\
    & - \mathbb{E}_{x^t\sim \mathcal{D}_t}\log\left[1-D(F(x^t))\right] \\
   \end{split}
   \label{eq:al}
\end{equation}
Here, $F$ is the feature extractor; $G$ is the classifier; $D$ is the domain discriminator. As suggested in the above two objectives, the domain discriminator aims to distinguish the samples of the source domain from those of the target domain. In this way, it can gradually diminish the domain shift and learn domain-invariant image features that are transferable across both source and target domains. And thus the classifier trained with merely the labeled samples from the source domain can be used to classify samples from both domains.

Plenty of works applied the above adversarial mechanism to domain adaptation task, but they mainly extract holistic features for domain adaptation and usually ignore local patterns that are more transferable and discriminative. With regard to the cross-domain FER task, these local features are more essential as this task requires a fine-grained and detailed understanding of the face images. To address these issues, we propose to integrate graph propagation networks with adversarial learning mechanisms to learn fine-grained and domain-invariant features. To this end, we extract several discriminative regions based on facial landmarks, and build an intra-domain graph to correlate holistic-local regions within each domain and an inter-domain graph to correlate these regions across different domains. We develop a class-aware two-stage updating mechanism to iteratively learn per-class statistical feature distribution for both holistic-local regions from both domains. Given an input image from one domain, we extract holistic-local features from corresponding regions to initialize graph nodes of this domain and apply the statistical feature distribution to initialize graph nodes of the other domain. Finally, we apply two stacked GCNs to propagate messages through the intra-domain graph to explore holistic-local feature interactions and transfer information across the inter-graph to enable holistic-local co-adaptation. An overall pipeline of the proposed AGRA framework is illustrated in Figure \ref{fig:framework}.

\subsection{Graph Construction}
In this part, we introduce the constructions of the intra-domain and inter-domain graphs. As suggested in previous works, local regions around some specified landmarks play essential roles for expression recognition. Thus, we extract the holistic face and further crop five local regions centered on left eye (\textit{le}), right eye (\textit{re}), nose (\textit{no}), left mouth corner (\textit{lm}), right mouth corner (\textit{rm}). We then build the two graphs $\mathcal{G}_{intra}=(\mathbf{V}, \mathbf{A}_{intra})$ and $\mathcal{G}_{inter}=(\mathbf{V}, \mathbf{A}_{inter})$. $\mathbf{V}=\{v^s_h, v^s_{le}, v^s_{re}, v^s_{no}, v^s_{lm}, v^s_{rm}, v^t_h, v^t_{le}, v^t_{re}, v^t_{no}, v^t_{lm}, v^t_{rm}\}$ is the node set denoting the holistic image and five local regions of source and target domains, and it is the same for both the two graphs. $\mathbf{A}_{intra}$ is the prior intra-domain adjacent matrix denoting the connections among nodes within each domain. It contains two type of connections, where the first type is global-to-local connection and and second type is local-to-local connections. $\mathbf{A}_{inter}$ is the prior inter-domain adjacent matrix denoting the connections between nodes from the different domains. Similarly, it contains three types of connections, i.e., global-to-global connection, global-to-local connection, and local-to-local connection. We use different values to denote different connections.

\subsection{Graph Representation Adaptation}
Once the two graphs are constructed, message propagations are performed through the intra-domain graph to explore holistic-local feature interactions with each domain and through the inter-domain graph to enable holistic-local feature co-adaptation. As suggested in previous works \cite{kipf2016semi}, graph convolutional network (GCN) \cite{kipf2016semi} can effectively update node features of graph-structured data by iteratively propagating node massages to the neighborhood nodes. In this work, we apply two stacked GCNs to propagate messages through the two graphs, respectively.

As discussed above, the graphs contain nodes of two domains. Given an input sample of one domain $d$ $(d\in\{s, t\})$, we extract the features of corresponding regions to initialize nodes of domain $d$. It is expected that these features can interact with feature distributions of the other domain, and thus the model can gradually diminish the domain shift. Besides, motivated by previous work \cite{wang2018stratified}, it is essential to integrate class information to enable finer-grained intra-class interaction and adaptation. To this end, we estimate the per-class statistical feature distributions of each domains i.e., $\bar\mu^s_{ck}$ and $\bar\mu^t_{ck}$ where $c\in \{0, 1, \dots, C-1\}$ is the class label and $k\in \{h, le, re, no, lm, rm\}$ is the node type. It is implemented by a class-aware two-stage updating mechanism as follows.

\subsubsection{Class-aware two-stage updating mechanism}
Here, we update the statistical distribution by epoch-level clustering that re-cluster the samples to obtain the distribution every $E$ epochs and iteration-level updating that updates the distribution every iteration. Specifically, we first extract features for all samples of both source and target datasets using the backbone network pre-trained using the labeled source samples. For each domain, we divide the samples into $C$ clusters using the K-means algorithm and compute the mean values for each cluster to obtain the initial statistical distribution, formulated as
\begin{equation}
   \begin{split}
    \bar{\mu}^s_{ck}&=\frac{1}{n^s_c}\sum_{i=1}^{n^s_c}f_k(x^s_{ci})\\
    \bar{\mu}^t_{ck}&=\frac{1}{n^t_c}\sum_{j=1}^{n^t_c}f_k(x^t_{ci})
   \end{split}
   \label{eq:average1}
\end{equation}
where $f_k(\cdot)$ is the feature extractor for region $k$; $n^{s/t}_c$ is the sample number of cluster $c$ of domain $s/t$; $x^{s/t}_{ci}$ is the $i$-th sample of cluster $c$. During training, we further use moving average to iteratively update these statistical distributions in a fine-grained manner. For each batch iteration, we compute the distances between each sample and the distributions of each cluster. These samples are grouped into the cluster with the smallest distance. Then, we compute the mean features (i.e., ${\mu}^s_{ck}$ and ${\mu}^t_{ck}$) over samples of the same cluster and update the statistical distribution by
\begin{equation}
   \begin{split}
    \bar{\mu}^s_{ck}&=(1-\alpha)\bar{\mu}^s_{ck}+\alpha {\mu}^s_{ck}\\
    \bar{\mu}^t_{ck}&=(1-\alpha)\bar{\mu}^t_{ck}+\alpha {\mu}^t_{ck}
   \end{split}
   \label{eq:average2}
\end{equation}
where $\alpha$ is a balance parameter, and it is set as 0.1 in our experiments. To avoid the distribution shift, this process is repeated by $E$ epochs. Then, we re-cluster the samples to obtain new distributions for each cluster according to Equation \ref{eq:average1}. The epoch-level re-clustering and iteration-level updating are iteratively performed along with the training process to obtain the final statistical distributions.

\subsubsection{Stacked graph convolution networks}
As discussed above, we use two stacked GCNs, in which one GCN propagates messages through the intra-domain graph to explore holistic-local feature interactions within each domain and another GCN transfers messages through the inter-domain GCN to enable holistic-local feature co-adaptation. In this part, we describe the two GCNs in detail.

Given an input sample $x^s_i$ from the source domain, we can extract features of the holistic image and the corresponding local regions to initialize the corresponding node of the source domain 
 \begin{equation}
   h_{intra, k}^{s,0}=f_{k}(x^s_i).
 \end{equation} 
Then, we compute the distance of this sample with the feature distributions of all clusters of the target domain, and obtain the cluster $c$ with smallest distance. Then, each node of the target domain is initialized by corresponding feature distribution
\begin{equation}
  h_{intra, k}^{t,0}=\bar{\mu}^t_{ck}.
\end{equation} 
The initial features are then re-arranged to obtain feature matrix $\mathbf{H}^0_{intra}\in \mathcal{R}^{n\times d^0_{intra}}$, where $n=12$ is the node number. Then, we perform graph covolution operation on the input feature matrix to iteratively  propagate and update node features, formulated as
\begin{equation}
   \mathbf{H}^l_{intra}=\sigma(\widehat{\mathbf{A}}_{intra}\mathbf{H}^{l-1}_{intra}\mathbf{W}^{l-1}_{intra}),
\end{equation} 
By stacking $L_{intra}$ graph convolution layers, the node messages are fully explored within the intra-domain graph and the feature matrix $\mathbf{H}_{intra}$ are obtained. This feature matrix is then used to initialize the nodes of inter-domain graph
\begin{equation}
  \mathbf{H}^0_{inter}=\mathbf{H}_{intra}.
\end{equation}
And the graph convolution operation is performed to iteratively update node features
\begin{equation}
   \mathbf{H}^l_{inter}=\sigma(\widehat{\mathbf{A}}_{inter}\mathbf{H}^{l-1}_{inter}\mathbf{W}^{l-1}_{inter}),
\end{equation}
Similarly, the graph convolution operation is repeated by $L_{inter}$ times and the final feature matrix $\mathbf{H}$ is generated. We concatenate the features of nodes from the source domain as the final feature $F(x^s_i)$, which is fed into the classifier to predict expression label and domain discriminator to estimate its domain. The two matrices $\widehat{\mathbf{A}}_{intra}$ and $\widehat{\mathbf{A}}_{inter}$ are initialized by the prior matrices $\mathbf{A}_{intra}$ and $\mathbf{A}_{inter}$ and jointly fine-tuned to learn better relationships during the training process.

Similarly, given a sample from the target domain, the nodes of the source domain are initialized by the corresponding extracted feature and those of the target domain are initialized by the corresponding statistical feature distributions. Then the same process is conducted to obtain the final feature $F(x^t_i)$. As it does not have expression label annotation, it is merely fed into the domain discriminator for domain estimation.

\subsection{Implementation Details}

\subsubsection{Network architecture}
We use ResNet50-variant \cite{he2016deep,zhao2019multi} that consists of four block layers as the backbone network to extract features. Given an input image of size $112 \times 112$, we can obtain feature maps of $28\times 28\times 128$ from the second layer and feature maps of size $7\times 7\times512$ from the fourth layer. For the holistic feature, we perform a convolution operation to obtain feature maps of size $7\times 7\times64$, which is followed by an average pooling layer to obtain a 64-dimension vector. For local features, we use MT-CNN \cite{zhang2016joint} to locate the landmarks and use feature maps from the second layer with a larger resolution. Specifically, we crop $7\times 7 \times 64$ feature maps center at the corresponding landmark and use similar convolution operation and average pooling to obtain a 64-dimension vector for each region.

The intra-domain GCN consists of two graph convolutional layers with output channels of 128 and 64, respectively. Thus, the sizes of parameter matrices $\mathbf{W}^{0}_{intra}$ and  $\mathbf{W}^{1}_{intra}$ are $64\times 128$ and $128\times 64$, respectively. The inter-domain GCN contains merely one graph convolutional layer and the output channel is also set as 64. The parameter matrix $\mathbf{W}^{0}_{inter}$ is with size of $64\times 64$. We perform ablative studies to analyze the effect of the layer number of the two GCNs and find setting them as 2 and 1 obtains the best results.

The classifier is simply implemented by a fully-connected layer that maps the 384-dimension (i.e., $64\times 6$) feature vector to seven scores that indicate the confidence of each expression label. Domain discriminator is implemented by two fully-connected layers with ReLU non-linear function, followed by another fully-connected layer to one score to indicate its domain.

\subsubsection{Training details}
The AGRA framework is trained with the objectives of equation \ref{eqn:al1} and \ref{eqn:al2} to optimize the feature extractor, classifier, and domain discriminator. Here, we follow previous domain adaptation works \cite{wen2019exploiting} to adopt a two-stage training process. We initialize the parameters of the backbone networks with those pre-trained on the MS-Celeb-1M \cite{guo2016ms} dataset and the parameters of the newly-added layers with the Xavier algorithm \cite{glorot2010understanding}. In the first stage, we train the feature extractor and classifier with the cross-entropy loss using SGD with an initial learning rate of 0.0001, a momentum of 0.9, weight decay of 0.0005. It is trained by around 15 epochs. In the second stage, we use the objective loss in equation \ref{eqn:al1} to train the domain discriminator and the objective loss in equation \ref{eqn:al2} to fine-tune the feature extractor and the classifier. It is also trained using SGD with the same momentum and weight decay as the first stage. The learning rate for feature extractor and the source classifier is initialized 0.0001, and it is divided by 10 after about 10 epoch. As the domain discriminator is trained from scratch, we initialize 0.001 and divide it by 10 when error saturates.

\subsubsection{Inference details}
Given an input image, we extract holistic and local images to initialize the corresponding nodes of the target domain. Then, we compute the distances between the given image and all the per-class feature distributions of the source domain. We select the feature distributions with smallest distance to initialize the nodes of the source domain. After GCN message propagation, we can obtain its feature and feed it into the classifier to predict the final score vector.

\section{Experiments}

\subsection{Datasets}

\begin{table*}[!t]
\centering
\begin{tabular}{c|c|c|cccccc}
\hline
\centering  Methods & Source sets & Backbone & CK+ & JAFFE & SFEW2.0 &  FER2013 &  ExpW & Mean\\
\hline
\hline
Da et al. \cite{da2015effects} & BOSPHORUS & HOG \& Gabor filters & 57.60 & 36.2 & - & - & - & -\\
Hasani et al. \cite{hasani2017facial} & MMI\&FERA\&DISFA & Inception-ResNet & 67.52 & - & - & - & -& -\\
Hasani et al. \cite{hasani2017spatio} & MMI\&FERA & Inception-ResNet & 73.91 & - & - & - & -& -\\
Zavarez et al. \cite{zavarez2017cross} & six datasets & VGG-Net & 88.58 & 44.32 & - & - & -& -\\
Mollahosseini et al. \cite{mollahosseini2016going} &  six datasets& Inception &  64.20 & - &39.80 & 34.00 & -& -\\
Liu et al. \cite{liu2015inspired} & CK+ & Manually-designed Net & - & - & 29.43 & - & - & -\\
DETN \cite{li2018deep} & RAF-DB & Manually-designed Net & 78.83 & 57.75 & 47.55 & 52.37  & -& -\\
ECAN \cite{li2020deeper} & RAF-DB 2.0 & VGG-Net & 86.49 & 61.94 & 54.34 & 58.21 & -& -\\
\hline
\hline
CADA \cite{long2018conditional} & RAF-DB &ResNet-50 & 72.09 & 52.11 & 53.44 & 57.61 & 63.15 & 59.68\\
SAFN \cite{xu2019larger} & RAF-DB &ResNet-50& 75.97 & 61.03 & 52.98 & 55.64 & 64.91 & 62.11 \\
SWD \cite{lee2019sliced} & RAF-DB &ResNet-50&75.19 & 54.93 & 52.06 & 55.84 &  68.35 & 61.27 \\
LPL \cite{li2017reliable} & RAF-DB &ResNet-50& 74.42 & 53.05 & 48.85 & 55.89 & 66.90 & 59.82 \\
DETN \cite{li2018deepemotion} &RAF-DB &ResNet-50& 78.22 & 55.89 & 49.40 & 52.29 & 47.58 & 56.68 \\
ECAN \cite{li2020deeper} &RAF-DB &ResNet-50& 79.77 & 57.28 & 52.29 & 56.46 & 47.37 & 58.63 \\
\hline
\hline
Ours & RAF-DB & ResNet-50 & \textbf{85.27} & \textbf{61.50} & \textbf{56.43} & \textbf{58.95}  & \textbf{68.50} & \textbf{66.13} \\
\hline
\end{tabular}
\vspace{2pt}
\caption{Accuracies of our proposed framework with current leading methods on the CK+, JAFFE, SFEW2.0, FER2013, and ExpW datasets. The results of the upper part are taken from the corresponding paper, and the results of the bottom part are generated by our implementation with exactly the ResNet-50 as backbone and RAF-DB as the source dataset. Work \cite{mollahosseini2016going} selects one dataset (i.e., CK+, SFEW2.0 or FER2013) from the CK+, MultiPIE, MMI, DISFA, FERA, SFEW2.0, and FER2013 as the target domain, and use the rest six datasets as the source domain; Work \cite{zavarez2017cross} selects one dataset (i.e., CK+ or JAFFE) from the CK+, JAFFE, MMI, RaFD, KDEF, BU3DFE and ARFace as the target domain, and the rest six datasets as the source domain.}
\vspace{-20pt}
\label{table:result}
\end{table*}

\noindent\textbf{CK+ \cite{lucey2010extended} }is a lab-controlled dataset that mostly used for evaluating FER. It contains 593 videos from 123 subjects, among which 309 sequences are labeled with six basic expressions based on the Facial Action Coding System (FACS). We follow previous work \cite{li2018deep} to select the three frames with peak formation from each sequence and the first frame (neutral face) of each sequence, resulting in 1,236 images for evaluation.


\noindent\textbf{JAFFE \cite{lyons1998coding}} is another lab-controlled dataset that contains 213 images from 10 Japanese females. Each person has about 3-4 images annotated with one of the six basic expressions and 1 image that is annotated with the neutral expression. This dataset mainly covers the Asian person and could be used for cross-race evaluation.

\noindent\textbf{FER2013 \cite{goodfellow2015challenges} }is a large-scale uncontrolled dataset that automatically collected by Google Image Search API. It contains 35,887 images with a size of 48$\times$48 pixels and each image is annotated with the seven basic expressions. The dataset is further divided into a training set of 28,709 images, a validation set of 3,589 images, and a test set of 3,589 images.

\noindent\textbf{SFEW2.0 \cite{dhall2011static} }is also an in-the-wild dataset collected from different films with spontaneous expressions, various head pose, age range, occlusions and illuminations. This dataset is divided into training, validation, and test datasets, with 958, 436, and 372 samples, respectively.

\noindent\textbf{ExpW \cite{zhang2018facial} }is an in-the-wild dataset and its images are downloaded from Google image search. This dataset contain 91,793 face images and each image is manually annotated with one of the seven expressions.

\noindent\textbf{RAF-DB \cite{li2018reliable} }contains 29,672 highly diverse facial images from thousands of individuals that are also collected from Internet. Among these, 15,339 images are annotated with seven basic expressions, which is divided into 12,271 training samples and 3,068 testing samples for evaluation. As this dataset is large in scale, and it serves well as the source dataset for cross-domain FER \cite{li2018deep}. We also select this dataset as our source domain and evaluate on the five above-mentioned datasets.


\subsection{Comparisons with state of the art}
As shown in Table \ref{table:result}, our approach achieves very competing performance on all datasets compared with current leading methods. For example, our approach outperforms all current methods on the SFEW2.0 and FER2013 datasets, while achieving comparable results with the best-performing ECAN \cite{li2020deeper} on the JAFFE dataset. 

However, current state-of-the-art methods use different backbone networks for image feature extraction and different datasets as the source domain. For example, works \cite{zhang2015facial,mayer2014cross} use MMI as the source dataset while works \cite{hasani2017facial,hasani2017spatio} further gather two datasets (i.e., MMI + JAFFE) as the source domain. These inconsistencies make the comparisons unfair and it is difficult to tell the effectiveness of each work. In addition, their backbone networks are also different, leading to extracted features with different discriminative abilities. This makes the comparisons even more unfair. To address these issues, we implement three best-performing methods according to the corresponding papers i.e., Locality Preserving Loss (LPL) \cite{li2017reliable}, Deep Emotion Transfer Network (DETN) \cite{li2018deepemotion}, and Emotion-Conditional Adaption Network (ECAN) \cite{li2020deeper}. To ensure fair comparisons, we use the same backbone network (i.e., ResNet-50) and source dataset (i.e., RAF-DB). Besides, there exist many general domain adaptation works, and we also apply some best-performing ones to cross-domain FER. To this end, we consider several newly-published works, i.e., Conditional Adversarial Domain Adaptation (CADA) \cite{long2018conditional}, Stepwise Adaptive Feature Norm (SAFN) \cite{xu2019larger} and Sliced Wasserstein Discrepancy (SWD) \cite{lee2019sliced}, and use the codes released by the authors for implementation. For fair comparisons, we also replace the backbone with ResNet-50 and the source dataset with RAF-DB. We evaluate the performance on CK+ \cite{lucey2010extended}, JAFFE \cite{lyons1998coding}, SFEW2.0 \cite{dhall2011static}, FER2013 \cite{goodfellow2015challenges}, and ExpW \cite{zhang2018facial}, as stated in the following. 

The results are presented in Table \ref{table:result}. When using the same backbone network and source dataset, the proposed AGRA consistently outperforms all current methods on all the datasets. Specifically, our AGRA approach obtains the accuracies of 85.27\%, 61.50\%, 56.43\%, 58.95\%, 68.50\% on the CK+, JAFFE, SFEW2.0, FER2013, ExpW, outperforming current best-performing methods by 5.5\%, 0.47\%, 2.99\%, 1.34\%, and 0.15\%, respectively. For a comprehensive comparison, we average the accuracies of all target datasets to obtain the mean accuracy. As shown, we find that the cross-domain FER methods like DETN and ECAN perform well for the lab-controlled CK+ dataset, but it performs quite poor for the more challenging in-the-wild (i.e., SFEW2.0, FER2013, and ExpW) and cross-race (JAFFE) settings. In contrast, general domain adaptation methods further propose to align features across different domains and can deal better with such challenging cases. For example, SAFN proposes to progressively adapt the feature norms of two domain, which achieves a mean accuracy of 62.11\%. Different from all these methods, the proposed AGRA approach integrates the graph propagation network with the adversarial mechanism for holistic-local feature co-adaptation across different domains to learn more domain-invariant and discriminative features, leading to obvious performance improvement. In particular, it achieves the mean accuracy of 66.13\%, with an improvement of 4.02\%.

\subsection{Ablative Study}
In this subsection, we conduct ablative studies to discuss and analyze the actual contribution of each component and give a more thorough understanding of the framework. To ensure fair comparison and evaluation, the experiments are conducted with the same ResNet-50 as the backbone and RAF-DB dataset as the source domain. We eliminate this information for simple illustrations.

\begin{table}[htp]
\centering
\small
\begin{tabular}{p{1.6cm}|p{0.7cm}p{0.7cm}p{0.7cm}p{0.7cm}p{0.7cm}p{0.7cm}}
\hline
\centering  Methods & CK+ & JAFFE & SFEW2.0 &  FER2013 &  ExpW & Mean\\
\hline
\hline
\centering Ours HF  & 72.09 & 52.11 & 53.44 & 57.61 & 63.15 & 59.68\\
\centering Ours HLF & 72.09 & 56.34 & 50.23 & 57.30 & 64.00 & 59.99\\
\centering Ours  & \textbf{85.27} & \textbf{61.50} & \textbf{56.43} & \textbf{58.95}  & \textbf{68.50} & \textbf{66.13} \\
\hline
\end{tabular}
\caption{Accuracies of our approach using holistic features (HF), concatenating holistic-local features (HLF) and ours for adaptation on the the CK+, JAFFE, SFEW2.0, FER2013, and ExpW datasets.}
\vspace{-25pt}
\label{table:result-hlf}
\end{table} 

\begin{figure*}[!t]
\centering
{
\includegraphics[width=0.95\linewidth]{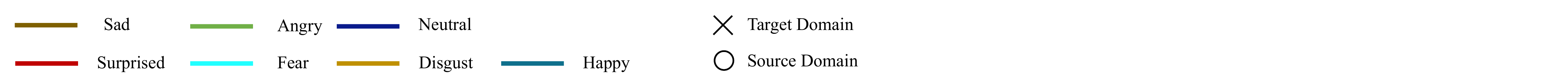}}
{
\label{fig:subfig1_1} 
\includegraphics[width=0.19\linewidth]{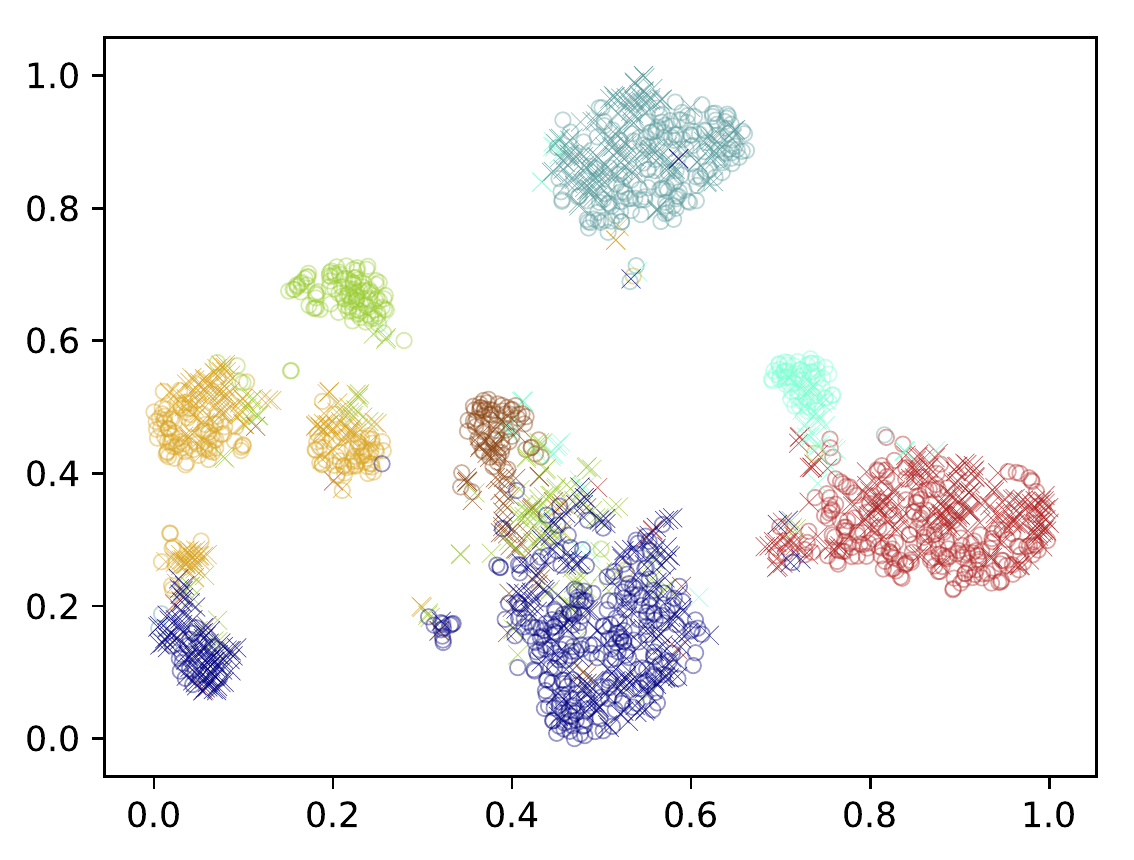}}
{
\label{fig:subfig1_2} 
\includegraphics[width=0.19\linewidth]{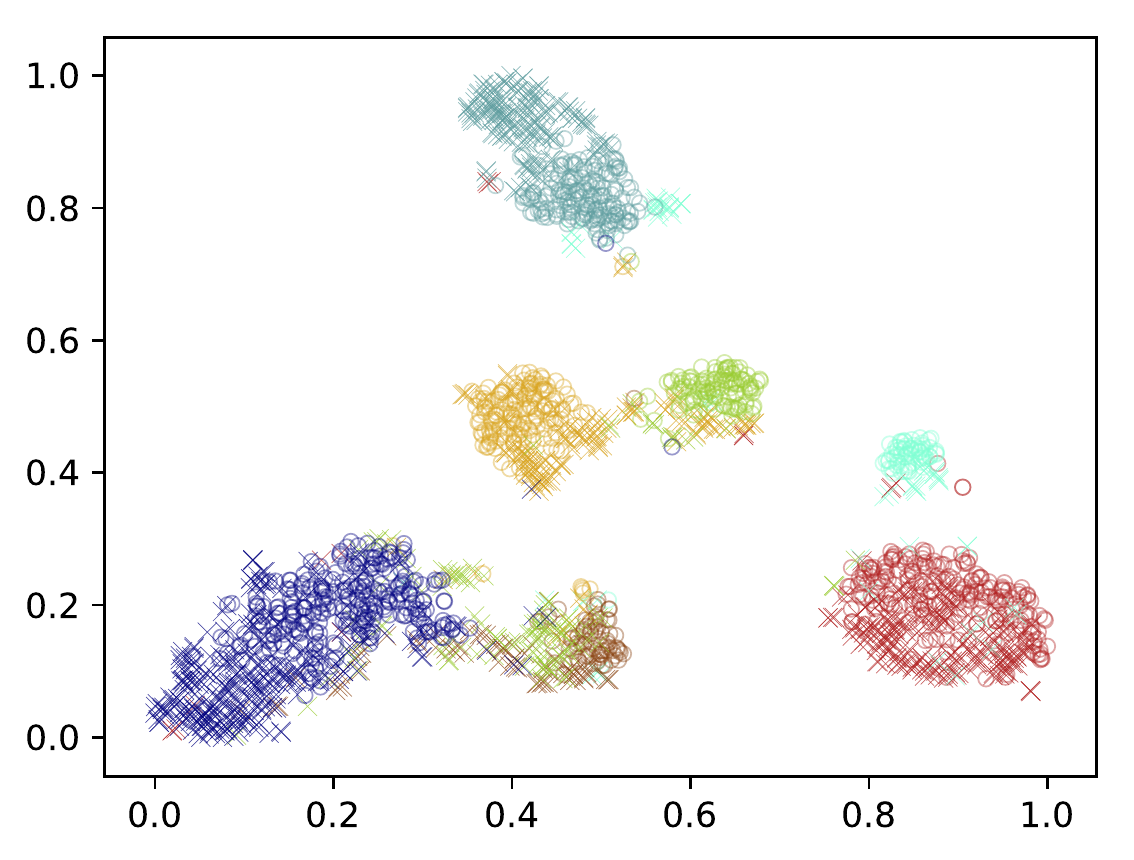}}
{
\label{fig:subfig1_3} 
\includegraphics[width=0.19\linewidth]{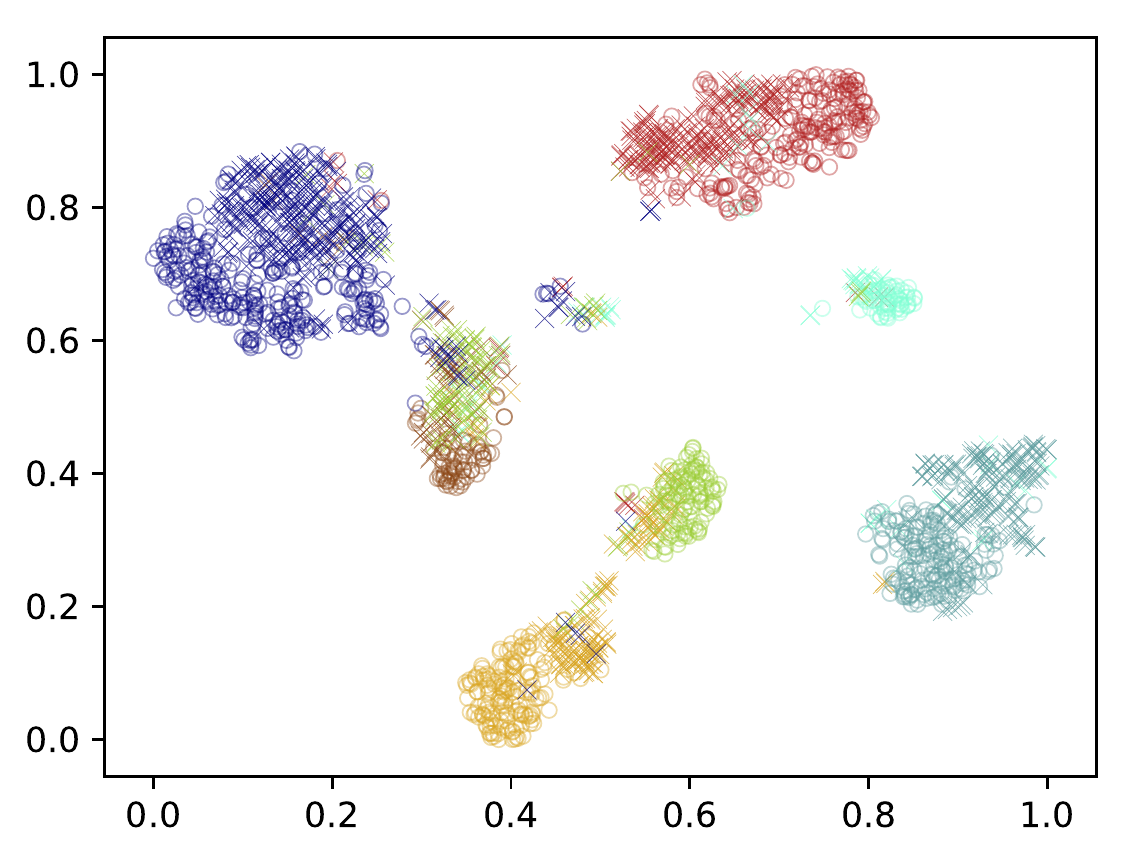}}
{
\label{fig:subfig1_4} 
\includegraphics[width=0.19\linewidth]{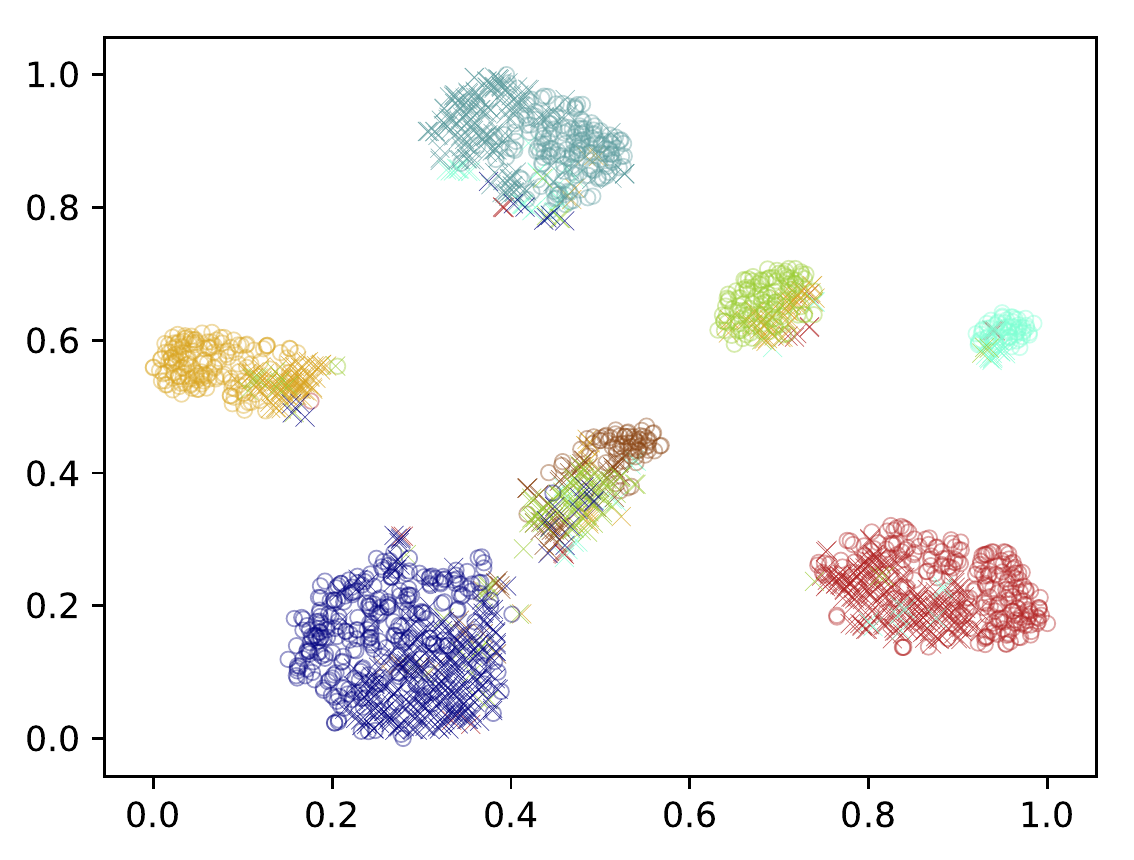}}
\vspace{-20pt}
{
\label{fig:subfig1_5} 
\includegraphics[width=0.19\linewidth]{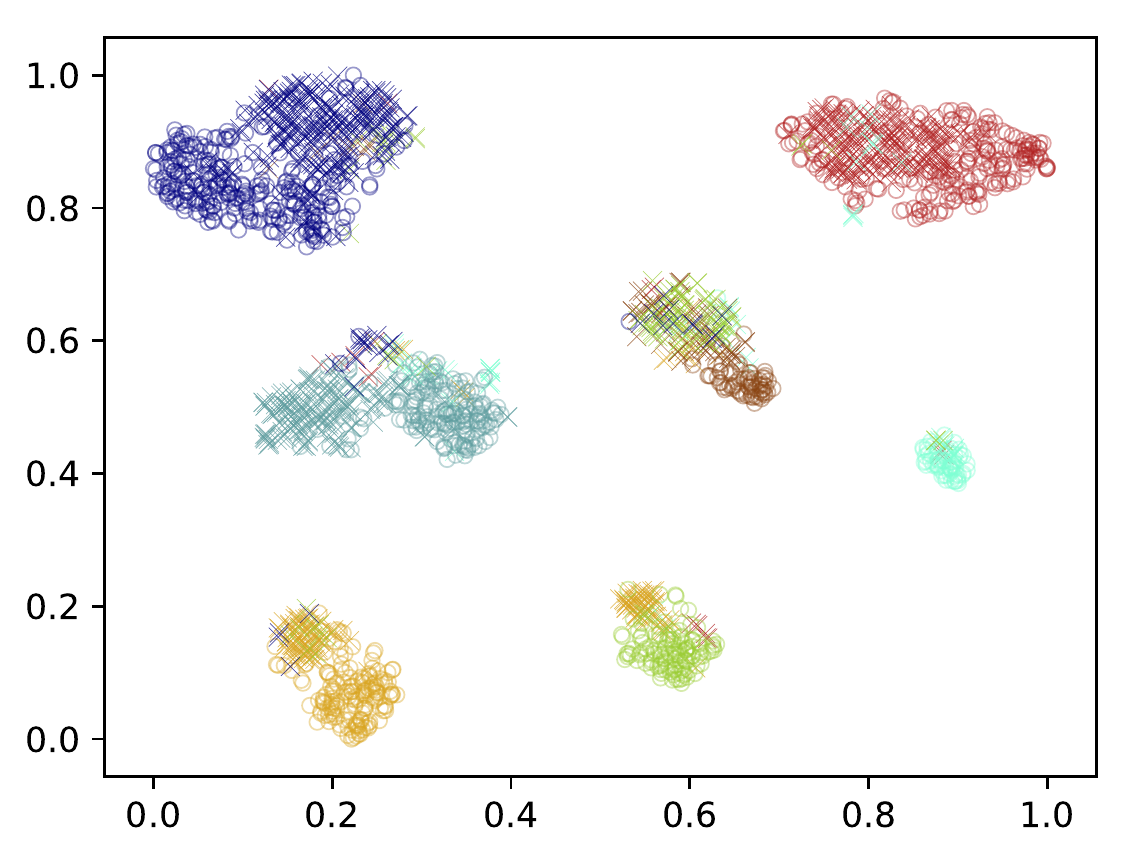}}
{
\label{fig:subfig2_1} 
\includegraphics[width=0.19\linewidth]{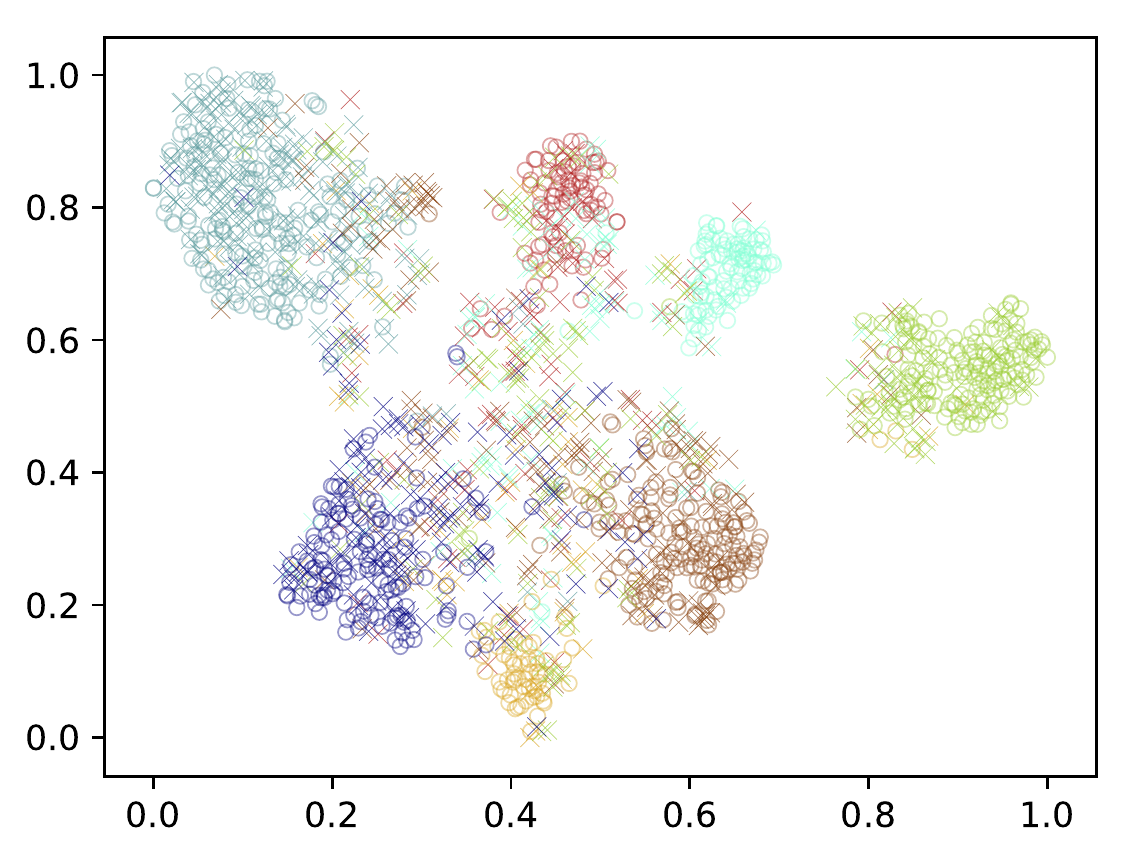}}
{
\label{fig:subfig2_2} 
\includegraphics[width=0.19\linewidth]{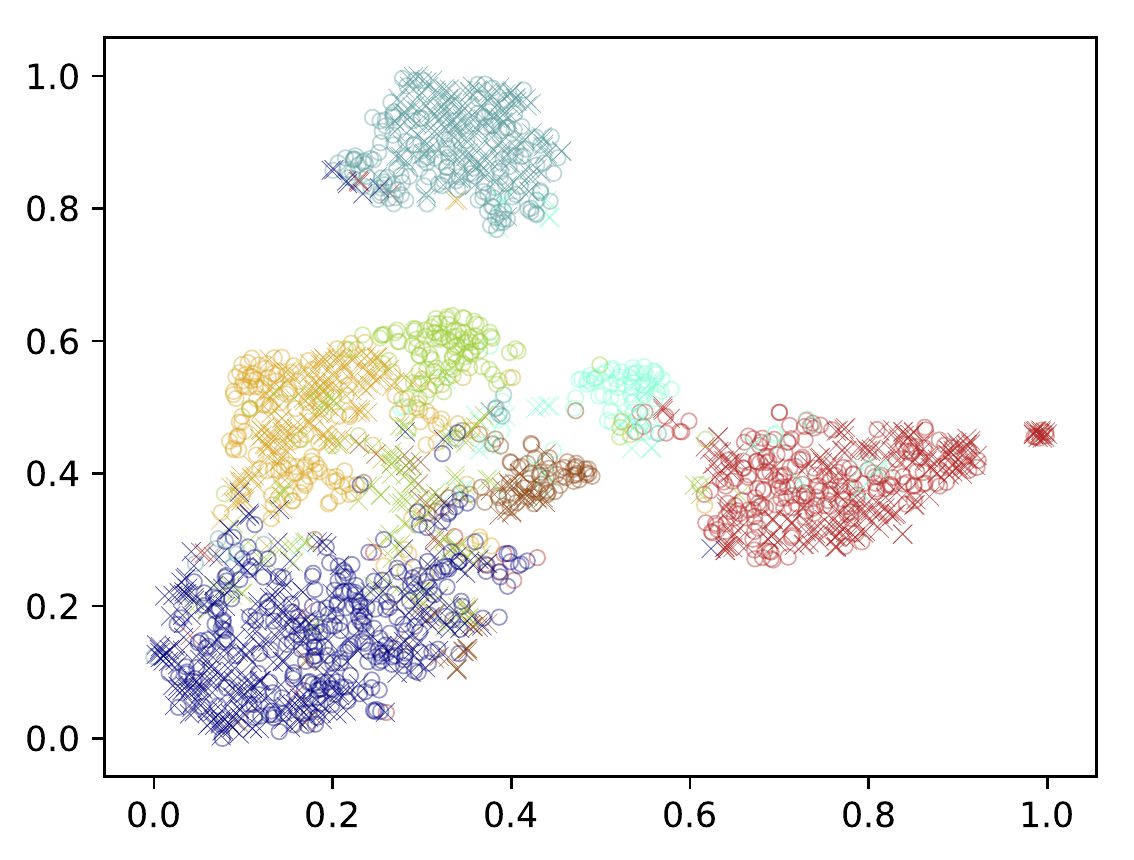}}
{
\label{fig:subfig2_3} 
\includegraphics[width=0.19\linewidth]{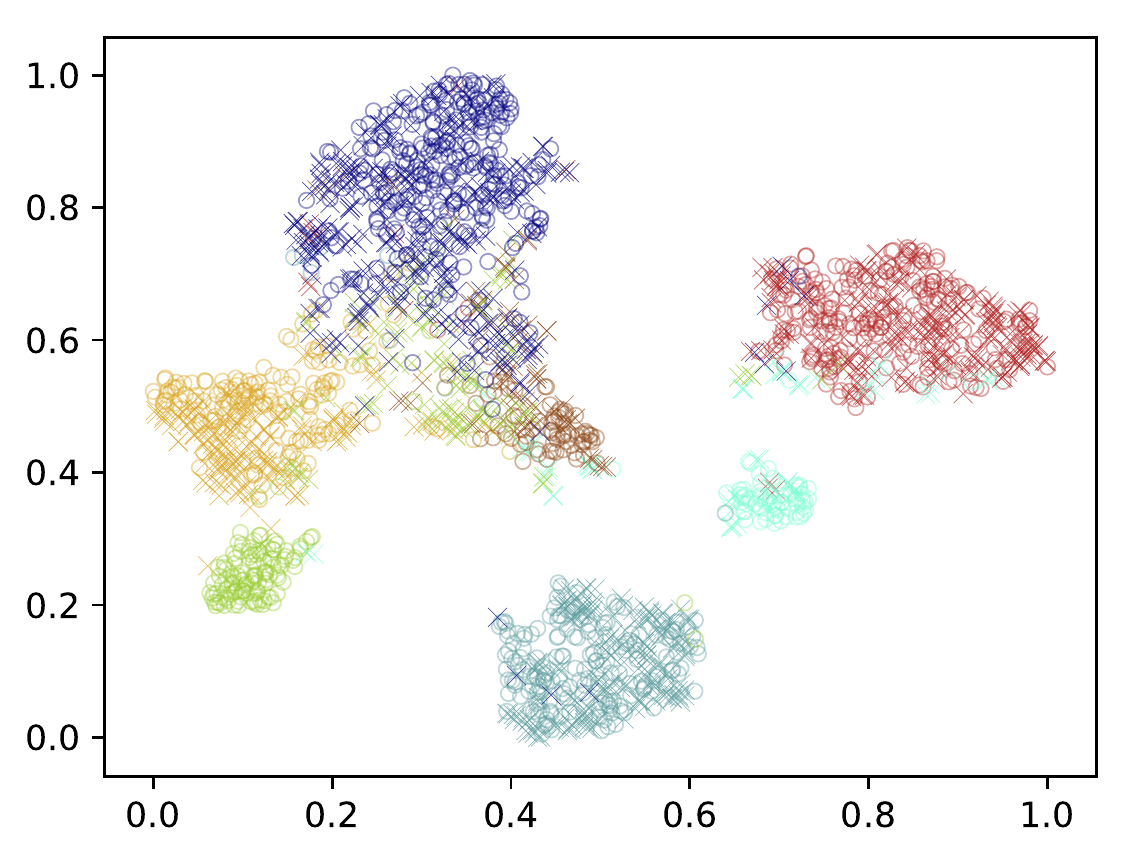}}
{
\label{fig:subfig2_4} 
\includegraphics[width=0.19\linewidth]{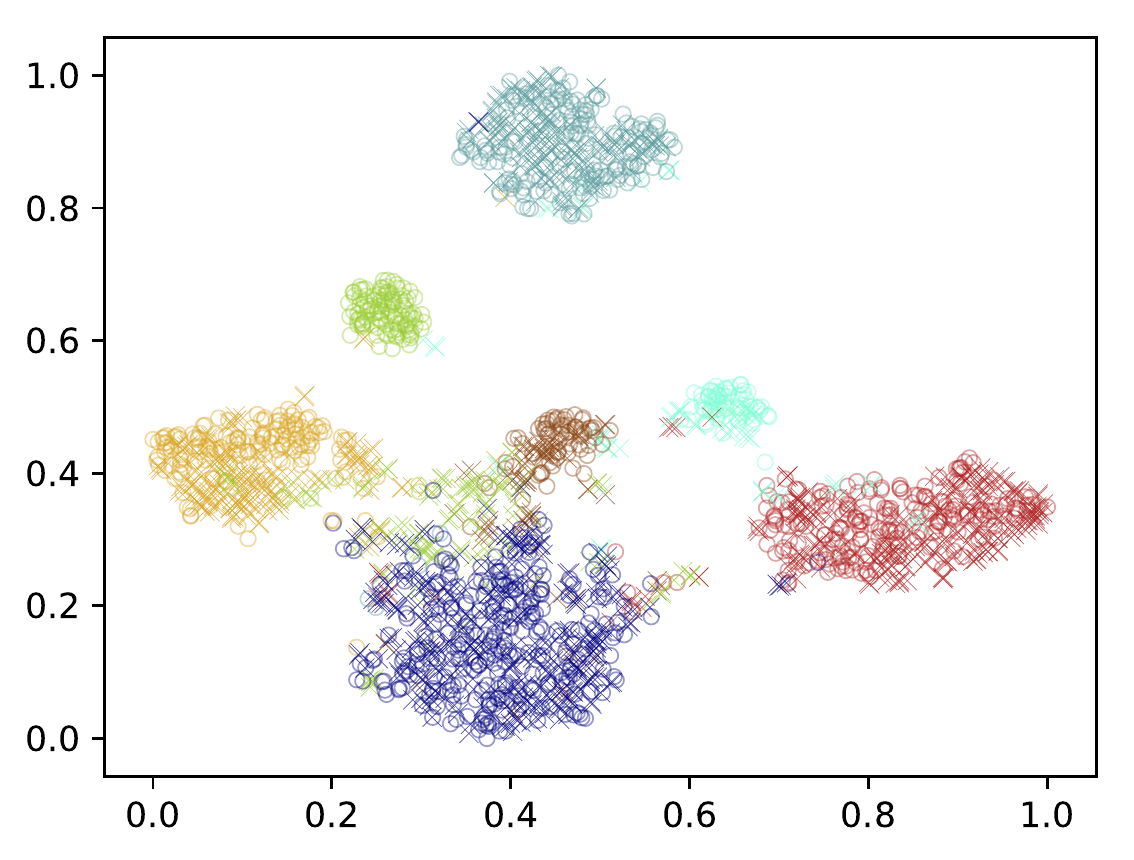}}
{
\label{fig:subfig2_5} 
\includegraphics[width=0.19\linewidth]{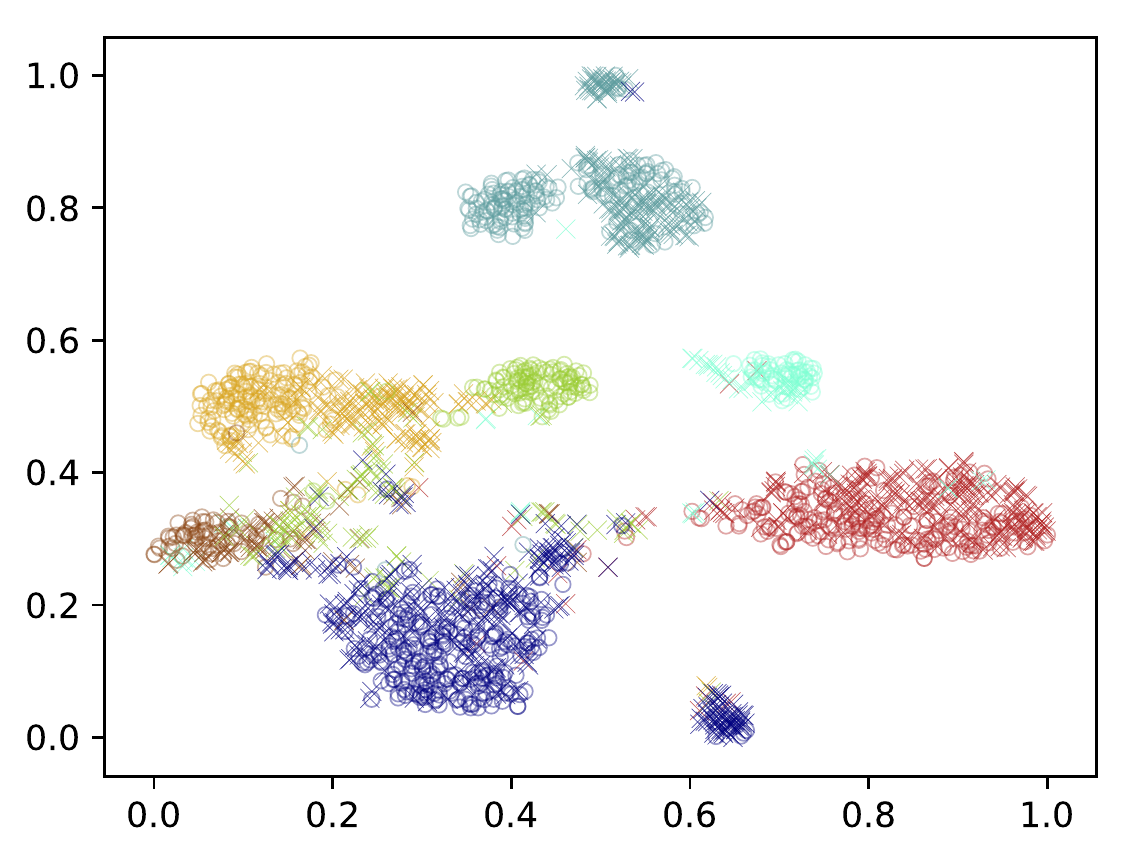}}
\vspace{-10pt}
\caption{Illustration of feature distribution learned by our proposed approach at epoch 0, 5, 10, 15, and 20 (from left to right) on the CK+ (upper) and SFEW2.0 (bottom) datasets.}
\label{fig:visualization}
\end{figure*}

\subsubsection{Analysis of holistic-local feature co-adaptation}
\label{sec:hlf}
The core contribution of the proposed framework is the holistic-local feature co-adaptation module that jointly learns domain-invariant holistic-local features. To analyze its contribution, we remove this module while keeping others unchanged. Thus, it merely uses holistic features for adaptation (namely Ours HF). As shown in Table \ref{table:result-hlf}, removing this module leads to obvious performance drop on all datasets. Specifically, the accuracies drop from 85.27\% to 72.09\% on CK+, from 61.50\% to 52.11\%, from 56.43\% to 53.44\%, from 68.50\% to 63.15\% on the five dataset, respectively. The mean accuracy drops from 66.13\% to 59.68\%, with a decreasing of 6.45\%. These obvious performance drops well demonstrate the contribution of the co-adaptation module for cross-domain FER. It is also key that we introduce two stacked GCN holistic-local feature co-adaptation. To verify its contribution, we remove the two GCN and simply concatenate holistic-local features for adaptation (namely Ours HLF). The results are also presented in Table \ref{table:result-hlf}. We find that concatenating local features can improve the performance, e.g., an improvement of 0.31\% in mean accuracy. However, it still performs much inferior to our AGRA approach on all five datasets, dropping the mean accuracy by 6.14\%.

\begin{table}[htp]
\centering
\small
\begin{tabular}{p{2cm}|p{0.7cm}p{0.7cm}p{0.7cm}p{0.7cm}p{0.7cm}p{0.7cm}}
\hline
\centering  Methods & CK+ & JAFFE & SFEW2.0 &  FER2013 &  ExpW & Mean\\
\hline
\hline
\centering Ours intra-GCN & 77.52 & \textbf{61.97} & 55.28 & 57.95 & 66.99  & 63.94\\
\centering Ours inter-GCN & 77.52 & 57.75 & 49.77 & 55.64 & 66.00 & 61.34\\
\centering Ours single GCN & 74.42 & 56.34 & 52.06 & 57.33 & 67.30 & 61.49\\
\centering Ours  & \textbf{85.27} & 61.50 & \textbf{56.43} & \textbf{58.95}  & \textbf{68.50} & \textbf{66.13} \\
\hline
\end{tabular}
\caption{Accuracies of our approach using merely the intra-domain GCN (Ours intra-GCN), using merely the inter-domain GCN (Ours inter-GCN), using merely one GCN (Ours single GCN), and Ours on the CK+, JAFFE, SFEW2.0, FER2013, and ExpW datasets.}
\vspace{-20pt}
\label{table:result-gcn}
\end{table} 

Note that we use two stacked GCN, in which an intra-domain GCN propagate messages within each domain to capture holistic-local feature interaction and an inter-domain GCN transfer messages across different domains to ensure domain adaptation. To demonstrate the effectiveness of this point, we conduct an experiment that uses one single GCN for message propagation within and across the source and target domains. As shown in Table \ref{table:result-gcn}, we find dramatic performance drops on all the datasets, e.g., decreasing the mean accuracy by 4.64\%. This is mainly because message propagations within each domain and across different domains are different, and using merely one GCN cannot model two types of propagation well. To further analyze the actual contribution of each GCN, we conduct two more experiments. The first experiment removes the inter-domain GCN and merely performs message propagation within each domain, while the second one removes the intra-domain GCN, and message propagation is merely carried out across different domains. We find that both two experiments show obvious performance drops, i.e., decreasing the mean accuracy by 2.19\% if removing the inter-domain GCN and by 4.79\% if removing the intra-domain GCN, as shown in Table \ref{table:result-gcn}.

\begin{table}[htp]
\centering
\small
\begin{tabular}{p{1.6cm}|p{0.7cm}p{0.7cm}p{0.7cm}p{0.7cm}p{0.7cm}p{0.7cm}}
\hline
\centering  Methods & CK+ & JAFFE & SFEW2.0 &  FER2013 &  ExpW & Mean\\
\hline
\hline
\centering Ours mean & 82.95 & 52.58 & 55.96 & 58.45 & 65.23 & 63.03 \\
\centering Ours iter  & 82.17 & 58.28 & 52.98 & 56.40 & 68.32 & 63.63 \\
\centering Ours epoch & 80.62 & 56.81 & 53.67 & 55.58 & 66.59 & 62.65 \\
\centering Ours  & \textbf{85.27} & \textbf{61.50} & \textbf{56.43} & \textbf{58.95}  & \textbf{68.50} & \textbf{66.13} \\
\hline
\end{tabular}
\caption{Accuracies of our approach with mean statistical distribution (Ours mean), per-class statistical distribution updated every each iteration (Ours iter), per-class statistical distribution updated every ten epochs (Ours epoch), on the CK+, JAFFE, SFEW2.0, FER2013, and ExpW datasets.}
\label{table:result-statistical-distribution}
\vspace{-25pt}
\end{table} 


\subsubsection{Analysis of the per-class statistical distribution}
To ensure meaningful initializations for nodes of each domain when the input image comes from the other domain, we learn the per-class statistical feature distribution. Here, we first illustrate the feature distributions of samples from the lab-controlled CK+ and in-the-wild SFEW2.0 datasets during different training stages. As shown in Figure \ref{fig:visualization}, it can be observed that the proposed model can gather the samples of the same category and from different domains together, which suggests that it can learn discriminative and domain-variant features. To quantitatively analyze its contribution, we learn the dataset-level statistical feature distributions and replace the per-class statistical feature distributions for node initialization. We find the mean accuracy drops from 66.13\% to 63.03\% as shown in Table \ref{table:result-statistical-distribution}. 

As stated above, we learn the per-class statistical distribution by updating every iteration and re-clustering every ten epochs. To analyze the effect of the updating mechanism, we conduct experiments that merely update every iteration or merely re-cluster every ten epochs, and present the results in Table \ref{table:result-statistical-distribution}. We find both experiments exhibit obvious performance drop, i.e., with mean accuracy drops by 2.50\% if using updating every iteration and by 3.48\% if using re-clustering every ten epochs.

\begin{table}[htp]
\centering
\small
\begin{tabular}{p{1.2cm}|p{0.7cm}p{0.7cm}p{0.7cm}p{0.7cm}p{0.7cm}p{0.7cm}}
\hline
\centering  Methods & CK+ & JAFFE & SFEW2.0 &  FER2013 &  ExpW & Mean\\
\hline
\hline
\centering Ours RM & 68.99 & 50.70 & 54.36 & 55.47 & 67.88 & 59.48 \\
\centering Ours OM & 79.07 & 57.28 & 53.90 & 57.07 & 66.71 & 62.81 \\
\centering Ours FM & 68.99 & 47.42 & 54.13 & 53.28 & 56.25 & 56.01 \\
\centering Ours  & \textbf{85.27} & \textbf{61.50} & \textbf{56.43} & \textbf{58.95}  & \textbf{68.50} & \textbf{66.13} \\
\hline
\end{tabular}
\vspace{2pt}
\caption{Accuracies of our approach where the matrices are initialized with randomly-initialized matrices (Ours RM), with all-one matrices (Ours OM), with fixed matrices (Ours FM), and ours on the CK+, JAFFE, SFEW2.0, FER2013, and ExpW datasets.}
\label{table:result-am}
\end{table}

\subsubsection{Analysis of adjacent matrix}
We initialize the two adjacent matrices of intra-domain and inter-domain graphs by manually defined connection, which can provide prior guidance to regularize message propagation. In this part, we replace the adjacent matrices with two randomly-initialized matrices (denoted as Ours RM) and with two all-ones matrices (denoted as Ours OM) to verify the effectiveness of this point. We present the results in Table \ref{table:result-am}. We observe both experiments show severe performance degradation on all datasets, i.e., degrading the mean accuracies by 6.65\% and 3.32\%. It is noteworthy that the experiment with randomly-initialized matrices exhibits more obvious performance degradation compared with the experiment with all-ones matrices. One possible reason is the randomly-initialized matrices may provide misleading guidance for message propagation, which further indicates the importance of the prior adjacent matrices.

To adjust the adjacent matrices to better guide message propagation, the adjacent matrices are also jointly fine-tuned during the training process. In this part, we verify its effectiveness by fixing the prior matrices training. We present the results in Table \ref{table:result-am}. The mean accuracy drops from 66.13\% to 56.01\%. This suggests jointly adjusting the adjacent matrices can learn dataset-specified matrices, which is crucial to promote cross-domain FER. 

\begin{table}[htp]
\centering
\small
\begin{tabular}{p{0.7cm}p{0.7cm}|p{0.7cm}p{0.7cm}p{0.7cm}p{0.7cm}p{0.7cm}p{0.7cm}}
\hline
\centering  $T_{intra}$ & $T_{inter}$ & CK+ & JAFFE & SFEW2.0 &  FER2013 & ExpW & Mean\\
\hline
\hline
\centering  1 & 1 & 75.19 & 52.11 & 55.28 & 57.22 & 67.32 & 61.42 \\
\centering  2 & 1 & \textbf{85.27} & \textbf{61.50} & \textbf{56.43} & \textbf{58.95}  & \textbf{68.50} & \textbf{66.13} \\
\centering  3 & 1 & 80.62 & 53.06 & 50.46 & 56.82 & 64.41 & 61.07\\
\hline
\hline
\centering  2 & 2 & 74.42 & 54.46 & 54.59 & 58.31 & 66.94 & 61.74 \\
\centering  2 & 3 & 79.07 & 49.77 & 51.61 & 56.85 & 67.14 & 60.89\\
\hline
\end{tabular}
\caption{Accuracies of our approach with different iteration numbers for the intra-domain GCN and inter-domain GCN on the CK+, JAFFE, SFEW2.0, FER2013, and ExpW datasets.}
\label{table:result-gcn-layer}
\vspace{-25pt}
\end{table}

\subsubsection{Analysis of the GCN iteration numbers}
As is known, increasing the layer number of GCN can promote deeper feature interaction, but it may lead to message smoothing and hurt their discriminative ability. Here, we present experimental studies to analyze the effect of iteration numbers (i.e., $T_{intra}$ and $T_{inter}$) of both GCNs on cross-domain FER. To this end, we first fix $T_{inter}$ as 1 and vary $T_{intra}$ from 1 to 3. As shown in Table \ref{table:result-gcn-layer}, it can boost the performance by increasing $T_{intra}$ from 1 to 2, but leads to performance drop when further increasing it to 3. Thus, we set the layer number of the intra-domain GCN as 2, and conduct an experiment that varies $T_{inter}$ from 1 to 3. We find setting $T_2$ as 1 obtains the best performance and increasing it suffers from performance degradation as depicted in Table \ref{table:result-gcn-layer}. Thus, we set $T_{intra}$ as 2 and $T_{inter}$ as 1 for all the experiments.

\section{Conclusion}
In this work, we develop a novel Adversarial Graph Representation Adaptation framework that integrates graph propagation mechanism with adversarial learning for holistic-local representation co-adaptation across different domains. We also explore learning per-class statistical distributions, which are used to initialize the graph nodes to help capture interactions between two domains. In the experiments, we perform extensive and fair experiments, where all methods use the same backbone network and source dataset, to demonstrate the effectiveness of the proposed framework.

\section{Acknowledgments}
This work was supported in part by the National Key Research and Development Program of China under Grant No. 2018YFC0830103, in part by National Natural Science Foundation of China (NSFC) under Grant No. 61876045 and 61836012, and in part by Zhujiang Science and Technology New Star Project of Guangzhou under Grant No. 201906010057.

\bibliographystyle{ACM-Reference-Format}
\bibliography{acmart}

%
%
%
%

\end{document}